%% file: main.tex
\documentclass[runningheads]{llncs}
\usepackage[T1]{fontenc}
\usepackage{graphicx}
\usepackage{amssymb}
\usepackage{marvosym}
\usepackage{bm}
\usepackage{amsmath}
\usepackage{diagbox}
\usepackage{subfigure}
\usepackage{xcolor}
\usepackage{multirow}
\usepackage{subcaption}
\usepackage{booktabs}
\usepackage{hyperref}
\usepackage{url}

\usepackage{color}

\begin{document}
\title{Cephalometric Landmark Detection across Ages with Prototypical Network}
\titlerunning{CeLDA: Cephalometric Landmark Detection across Ages}

\author{Han Wu \inst{1}\textsuperscript{*}, Chong Wang\inst{2}\textsuperscript{*}, Lanzhuju Mei\inst{1,3}, Tong Yang\inst{7}, Min Zhu\inst{6}, \\ Dinggang Shen\inst{1,4,5}, Zhiming Cui\inst{1}\textsuperscript{(\Letter)}}
\authorrunning{H Wu et al.}

%
\institute{School of Biomedical Engineering \& State Key Laboratory of Advanced Medical Materials and Devices, ShanghaiTech University, Shanghai, China\\
\email{cuizhm@shanghaitech.edu.cn}\and
Australian Institute for Machine Learning, The University of Adelaide\and
Lingang Laboratory, Shanghai, China\and
Shanghai United Imaging Intelligence Co. Ltd., Shanghai, China\and
Shanghai Clinical Research and Trial Center, Shanghai, China \and
Shanghai Ninth People's Hospital, Shanghai Jiao Tong University, Shanghai, China \and
Shanghai Linkedcare Information Technology Co., Ltd., Shanghai, China}

\newcommand{\chong}[1]{{\color{blue}[chong]: #1}}
\newcommand{\wh}[1]{{\color{orange}[wh]: #1}}

\maketitle
\begin{abstract}
Automated cephalometric landmark detection is crucial in real-world orthodontic diagnosis.
Current studies mainly focus on only adult subjects, neglecting the clinically crucial scenario presented by adolescents whose landmarks often exhibit significantly different appearances compared to adults.
Hence, an open question arises about how to develop a unified and effective detection algorithm across various age groups, including adolescents and adults. 
In this paper, we propose CeLDA, the first work for \textbf{Ce}phalometric \textbf{L}andmark \textbf{D}etection across \textbf{A}ges.
Our method leverages a prototypical network for landmark detection by comparing image features with landmark prototypes. 
To tackle the appearance discrepancy of landmarks between age groups, we design new strategies for CeLDA to improve prototype alignment and obtain a holistic estimation of landmark prototypes from a large set of training images.
Moreover, a novel prototype relation mining paradigm is introduced to exploit the anatomical relations between the landmark prototypes. 
Extensive experiments validate the superiority of CeLDA in detecting cephalometric landmarks on both adult and adolescent subjects. 
To our knowledge, this is the first effort toward developing a unified solution and dataset for cephalometric landmark detection across age groups. 
Our code and dataset will be made public on \href{https://github.com/ShanghaiTech-IMPACT/Cephalometric-Landmark-Detection-across-Ages-with-Prototypical-Network}{Github}.

\keywords{Cephalometric Landmark \and Prototypical Network \and Landmark Prototypes \and Relation Mining \and Prototype Alignment}
\end{abstract}

\section{Introduction}
Automatic and accurate detection of cephalometric landmarks holds significant importance in clinical practice, particularly for orthodontic diagnosis and therapy planning \cite{schwendicke2021deep}. 
With the remarkable achievements of deep learning~\cite{schmidhuber2015deep,wang2022bowelnet}, there are many learning-based efforts made for detecting cephalometric landmarks, i.e.,
regressing landmarks with deep convolutional neural networks~\cite{lee2017cephalometric,oh2020deep}, 
improving detection performance with two-stage networks~\cite{jiang2022cephalformer,lee2020automated,song2020automatic,zeng2021cascaded}, and
modelling landmark relationships with anatomical prior information~\cite{chen2022semi,li2020structured}. 

\begin{figure*}[t]
    \centering
    \includegraphics[width=0.9\textwidth, height=4.2 cm]{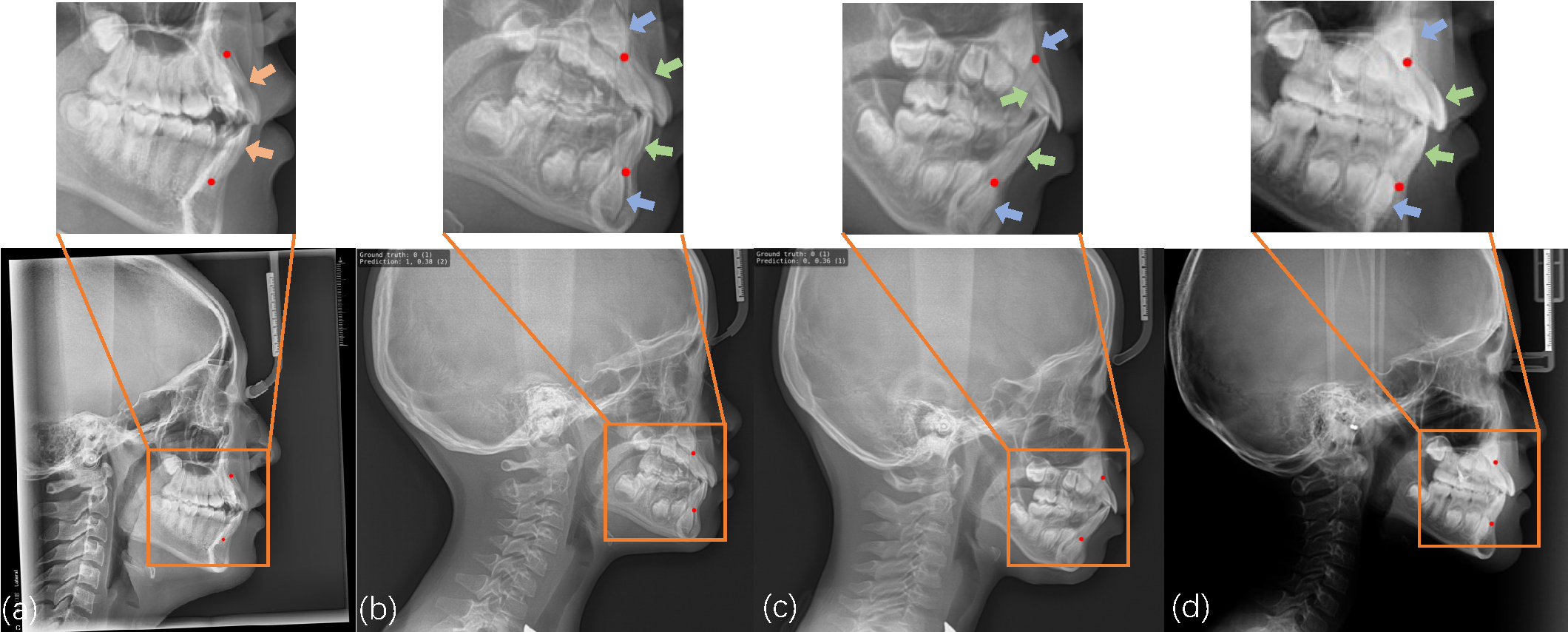}
    \caption{(a) An adult case, with regular anatomical structures and permanent teeth (orange arrow); (b,c,d) adolescent cases, with complicated anatomical changes due to unerupted teeth (blue arrow) and baby teeth (green arrow). 
    These changes on adolescent cases cause significant landmark shifts. Here we show only two landmarks (red points) out of ten for better visualization.}
    \label{fig:Difference}
\end{figure*}

Regardless of their encouraging performances, these existing approaches are mostly dedicated to detecting cephalometric landmarks on adult subjects,
which has clear skull bone and regular tooth arrangement shown in Fig.~\ref{fig:Difference}(a),
ignoring more challenging adolescent subjects that often have complicated morphological changes in anatomy due to the presence of unerupted and permanent teeth as in Fig.~\ref{fig:Difference}(b,c,d). 
Such changes are prone to cause significant shifts of the cephalometric landmarks~\cite{tanikawa2010automatic}. 
Very recently, Ceph-Net~\cite{yang2023ceph} targets the cephalometric landmark detection on adolescent cases and utilizes an attention-based stacked regression network to progressively refine detection results.
However, Ceph-Net considers only adolescent cases but the common adult cases are not included. 
To date, it remains unexplored and needs to be addressed to develop a unified and effective cephalometric landmark detection algorithm across different age groups, including both adolescent and adult cases.
Generally, the main obstacle in approaching such an algorithm comes from the landmark shifts across age groups, necessitating robust learning capabilities of the algorithm.

In this paper, we propose CeLDA for age-inclusive cephalometric landmark detection. 
Specifically, our CeLDA relies on a prototypical network to realize landmark detection by comparing image features with landmark prototypes.
To ensure robust prototypes against the landmark shifts from different age groups, 
we present new strategies for CeLDA to promote prototype alignment and obtain a holistic estimation of landmark prototypes from a large set of training samples.
Furthermore, a novel prototype relation mining paradigm is introduced to leverage anatomical relations among landmarks. 
Extensive experimental results illustrate that our CeLDA outperforms existing state-of-the-art (SOTA) approaches in detecting cephalometric landmarks on adolescent subjects, adult subjects, and both. 
To summarise, \textbf{ our major contributions are}: 
1) the first prototype-based approach for age-inclusive cephalometric landmark detection, where the holistic prototypes are obtained to improve the learning robustness and predictive performance;
2) a novel prototype relation mining paradigm to take advantage of crucial anatomical relationships between landmarks; 
3) a new comprehensive benchmark dataset for landmark detection that consists of cephalometric images from both adolescent and adult subjects.

\begin{figure*}[t]
    \centering
    \includegraphics[width=\textwidth]{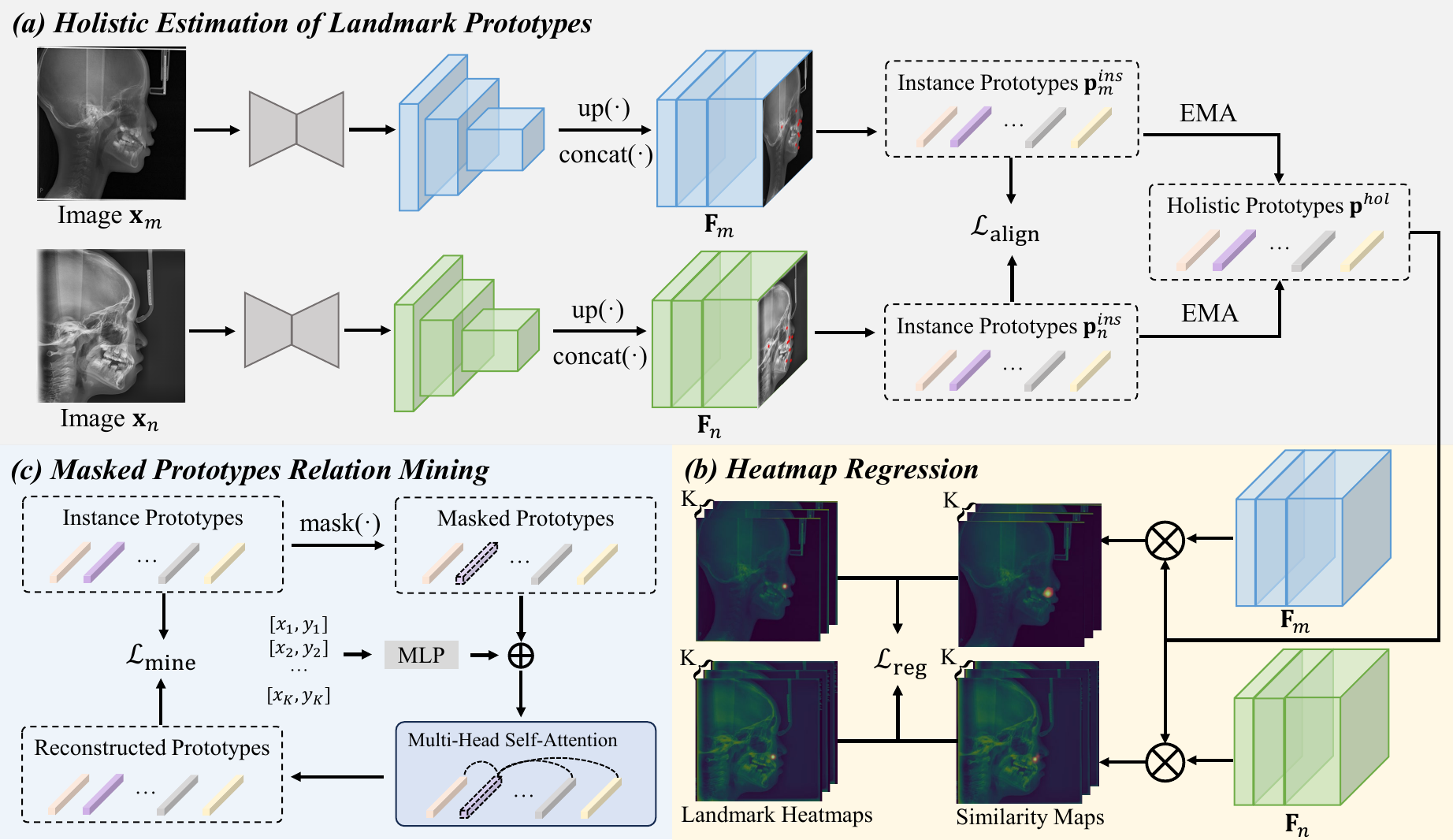}
    \caption{An overview of the proposed CeLDA method for cephalometric landmark detection, based on a set of holistic landmark prototypes. }
    \label{fig:pipeline}
\end{figure*}

\section{Methodology}

Our dataset $\mathcal{D}$ comprises image-label pairs represented by $(\mathbf{x}, \mathbf{y})$, where $\mathbf{x} \in \mathbb{R}^{H \times W}$ denotes a cephalometric image of size $H \times W$, 
and $\mathbf{y} \in \{0,1\}^{{K \times H \times W}}$ represents $K$ binary ground-truth landmark maps.
Each of the $K$ landmark maps only has one single annotated landmark point, i.e., $\sum \mathbf{y}_{k,:,:} = 1$. Following existing approaches~\cite{chen2022semi,he2021cephalometric,yueyuan2021swin,zhong2019attention}, we transform the sparsely-distributed landmark maps into $K$ landmark heatmaps $\mathbf{H}_k = \text{Gaussian}(\mathbf{y}_k) \in \mathbb{R}^{H \times W}$ for model training, using a Gaussian smoothing strategy as in~\cite{chen2022semi,zhu2021you}.

\subsection{Overview}
\label{sec:overview}
An overview of our proposed method is shown in Fig.~\ref{fig:pipeline}.
For an input cephalometric image $\mathbf{x}$, 
we employ a network backbone $f_{\theta}$, i.e., U-Net~\cite{ronneberger2015u}, to extract multi-level high-resolution feature maps $\{\mathbf{F}_1, \mathbf{F}_2, \mathbf{F}_3\}$, where 
$\mathbf{F}_1 \in \mathbb{R}^{\frac{H}{4} \times \frac{W}{4} \times D_1}$, 
$\mathbf{F}_2 \in \mathbb{R}^{\frac{H}{2} \times \frac{W}{2} \times D_2}$,
$\mathbf{F}_3 \in \mathbb{R}^{\frac{H}{1} \times \frac{W}{1} \times D_3}$.
In order to enable an accurate detection of the sparsely-distributed landmark, these feature maps are up-sampled to the original resolution of the input image and then concatenated into a composite feature map $\mathbf{F} = \text{concat}\left(\text{up}(\mathbf{F}_1), \text{up}(\mathbf{F}_2), \text{up}(\mathbf{F}_3)\right) \in \mathbb{R}^{H \times W \times D}$, where $D = (D_1 + D_2 + D_3)$ and $\text{up}(\cdot)$ denotes an up-sampling operation. 

In Fig.~\ref{fig:pipeline}(a), our CeLDA method leverages $K$ holistic prototypes $\mathcal{P}_{hol}=\{\mathbf{p}^{hol}_k\}_{k=1}^{K}$ 
that are estimated from a large set of training samples, as introduced in Section~\ref{sec:holistic}.
In $\mathcal{P}_{hol}$, each prototype $\mathbf{p}^{hol}_k \in \mathbb{R}^{1\times 1 \times D}$ corresponds to one landmark and captures robust landmark-representative features.
After that, we derive $K$ similarity maps, see Fig.~\ref{fig:pipeline}(b),
by calculating the dot-product between the feature maps $\mathbf{F}$ and each prototype $\mathbf{p}^{hol}_k$, which is formulated as:
\begin{equation}
    \mathbf{S}_k =\mathbf{p}^{hol}_k \cdot \mathbf{F} \in \mathbb{R}^{H \times W}.
    \label{eq:similarity}
\end{equation}
Finally, the detection prediction for the $k$-th landmark is obtained by selecting the location in $\mathbf{S}_k$ that has the highest similarity. 
For model training, the standard regression loss is utilized to supervise our CeLDA: 
\begin{equation}
    \mathcal{L}_{\text{reg}} = \frac{1}{K} \sum^{K}_{k} ||\mathbf{S}_k - \mathbf{H}_k||_2^2
\end{equation}
where $\mathbf{H}_k$ is the $k$-th ground-truth landmark heatmap.

In the following section, we elaborate how to estimate and obtain the prototypes $\mathcal{P}_{hol}$ for robust cephalometric landmark detection across age groups.

\subsection{Holistic Estimation of Landmark Prototypes}
\label{sec:holistic}

Prototypes have been studied for classification~\cite{snell2017prototypical} and segmentation~\cite{zhou2022rethinking} for a long time~\cite{wang2023interpretable}, 
where their essence is representing classes by prototypes to encode class-representative features. 
Making an analogy to our landmark detection task, it is natural to define prototypes to represent landmarks, i.e., capturing landmark-representative features. 
To achieve this, we propose to first create instance-level landmark prototypes $\mathcal{P}_{ins}=\{\mathbf{p}_k^{ins}\}_{k=1}^{K}$ for each individual training image $\mathbf{x}$, 
where $\mathbf{p}_k^{ins} \in \mathbb{R}^{1\times 1 \times D}$ and each of them is calculated as:
\begin{equation}
    \mathbf{p}_{k}^{ins} = \frac{\sum_{i, j} \mathbf{H}_{k}(i, j) \cdot \mathbf{F}(i, j)}{\sum_{i, j} \mathbf{H}_{k}(i, j)}
    \label{eq:localprototypes}
\end{equation}
where $i \in \{1, ..., H\}$ and $j \in \{1, ..., W\}$ are spatial indexes.
From Eq.~(\ref{eq:localprototypes}), the instance prototypes $\mathcal{P}_{ins}$ are generated by averaging the local contextual features around the landmark point. 
Although straightforward and easy to implement, one noticeable shortcoming of the instance-level prototypes is that they consider only individual-image information, 
which is insufficient to encapsulate the drastic appearance variations of the cephalometric landmarks, particularly for different age groups. 
To overcome this problem, we propose a new strategy to achieve a holistic estimation of the landmark prototypes. 
Specifically, inspired by the well-established exponential moving averaging (EMA) technique, we obtain the holistic prototypes $\mathcal{P}_{hol}$ in an on-the-fly fashion by exploiting a large set of training samples, as formulated below:
\begin{equation}
    \mathcal{P}_{hol}^{(t+1)} = \alpha \cdot \mathcal{P}_{hol}^{(t)} + (1-\alpha) \cdot \frac{1}{|\mathcal{B}|} \sum_{b=1}^{|\mathcal{B}|} \mathcal{P}_{ins}^{(t), b},
    \label{eq:ema}
\end{equation}
where $\mathcal{B}$ denotes a training mini-batch with size $|\mathcal{B}|$, 
$\alpha$ is a momentum update coefficient,
$t$ indicates the training iteration,
and $\mathcal{P}_{hol}$ is our holistic prototypes used for reliable detection of the landmarks, as described in Section~\ref{sec:overview}.
It is worth noting in Eq.~(\ref{eq:ema}) that during training our holistic prototypes are slowly progressing to take advantage of information from 
not only the current mini-batch but also historical prototypes.
Therefore, they will gradually gain a global picture of the whole training set, 
allowing a robust landmark detection from cephalometric images across ages, such as adolescent and adult stages.

\subsection{Cross-image Prototype Alignment}

According to Eq.~(\ref{eq:ema}), our holistic prototypes are obtained by accumulating instance-level prototypes during training, 
to increase the prototype robustness we also propose to encourage prototype alignment across individual images: 
\begin{equation}
    \mathcal{L}_{\text{align}} = \frac{1}{K} \sum_{k=1}^{K} ||\mathbf{p}_{m, k}^{ins} - \mathbf{p}_{n, k}^{ins} ||^2_2,
    \label{eq:alignment}
\end{equation}
where $\mathbf{p}_{m, k}^{ins}$ and $\mathbf{p}_{n, k}^{ins}$ denote the $k$-th instance-level prototypes for the image $\mathbf{x}_m$ and $\mathbf{x}_n$ within the mini-batch $\mathcal{B}$, respectively. 
Notice that Eq.~(\ref{eq:alignment}) is able to enforce prototype consistency for training samples from not only within the same age group but also across different age groups. 

\subsection{Masked Prototype Relation Mining}

As illustrated in Fig.~\ref{fig:Difference}, landmarks naturally have crucial anatomical relations within a cephalometric image \cite{yao2021one}.
Given that our CeLDA harnesses prototypes to represent landmarks, 
we further present a novel prototype relation mining paradigm to exploit the anatomical dependency between landmarks.

Motivated by the great success of masked modeling in language~\cite{devlin2018bert} and vision~\cite{he2022masked} applications, 
in this paper, we propose to mask the instance-level landmark prototypes. 
As demonstrated in Fig.~\ref{fig:pipeline}(c), after obtaining $K$ instance prototypes $\mathcal{P}_{ins}$ from a training image,
we randomly mask out a proportion of prototypes in $\mathcal{P}_{ins}$ and replace them with zero, 
where the landmark positional embeddings are introduced as a location indicator. 
The combination of masked prototypes and positional embeddings is processed by a multi-head self-attention (MSA) layer, 
which reconstructs the masked prototypes as follows: 
\begin{equation}
     \hat{\mathcal{P}}_{ins} = \text{MSA}\left(\text{mask}(\mathcal{P}_{ins}) \oplus \mathbf{E}_{pos}\right), \quad \mathbf{E}_{pos} = \text{MLP}(\mathbf{\bar{y}}),
     \label{eq:reconstructed}
\end{equation}
where $\hat{\mathcal{P}}_{ins}$ is the reconstructed prototypes, 
$\text{mask}(\cdot)$ is a mask-out operation to randomly exclude a ratio (denoted by $R$) of prototypes from $\mathcal{P}_{ins}$, 
$\oplus$ represents the element-wise summation,
and 
$\mathbf{E}_{pos}$ denotes the landmark positional embeddings, encoding the ground-truth landmark coordinates $\mathbf{\bar{y}} \in \mathbb{R}^{K \times 2}$ using a multi-layer perceptron (MLP), 
where the landmark coordinates $\mathbf{\bar{y}}$ can be easily derived from the ground-truth landmark maps $\mathbf{y}$.
The reconstructed prototypes are supervised by the original prototypes in $\mathcal{P}_{ins}$:
\begin{equation}
    \mathcal{L}_{\text{mine}} = \sum_{k=1}^{K} ||\hat{\mathbf{p}}_{k}^{ins} - \mathbf{p}_{k}^{ins} ||^2_2,
    \label{eq:mining}
\end{equation}
where $\hat{\mathbf{p}}_{k}^{ins}$ denotes a reconstructed prototype in $\hat{\mathcal{P}}_{ins}$, 
and $\mathbf{p}_{k}^{ins}$ is the corresponding raw prototype in $\mathcal{P}_{ins}$. 
Relying on Eq.~(\ref{eq:mining}), our CeLDA can make full use of the structural information regarding landmark relations 
during the process of learning the instance-level prototypes for each training sample, benefiting its understanding of the anatomical landmark dependency.

\subsection{Overall Training Objective}
The overall optimization objective of CeLDA is defined as:
\begin{equation}
    \mathcal{L}_{\text{total}} = \mathcal{L}_{\text{reg}} + \lambda_{1}\mathcal{L}_{\text{align}} + \lambda_{2}\mathcal{L}_{\text{mine}},
    \label{eq:totalloss}
\end{equation}
where $\lambda_{1}$ and $\lambda_{2}$ are hyper-parameters to control the weight of the loss terms.

\section{Experiments}

\subsection{Dataset and Evaluation Metric}

\subsubsection{Dataset:} 
For the task of cephalometric landmark detection across age groups, we collected a new benchmark dataset, named CephAdoAdu, with both adolescent and adult cases,
distinguishing it from existing datasets that solely consist of either adolescent or adult cases. 
CephAdoAdu has a total of 1000 (500 adult cases, 500 adolescent cases) cephalometric X-ray images, acquired from eight clinical centers. 
Every cephalometric image underwent manual annotations to mark 10 typical landmarks, by an experienced dental radiologist with over ten years of expertise.
Our new dataset has two advantages over existing ones: 
1) a more clinically practical coverage of subjects across different age groups; 
2) a larger number of annotated images, ensuring a comprehensive and faithful model evaluation. 
The whole dataset is randomly divided into training set (400 images), validation set (300 images), and testing set (300 images).
Notice that our data split is evenly performed in terms of the adult and adolescent cases.

\subsubsection{Evaluation Metric:} 
Following previous studies~\cite{wang2015evaluation,wang2016benchmark}, we evaluate the model performance with the two commonly-used metrics: 
1) Mean Radial Error (MRE) computes the average Euclidean distance between the predicted and ground-truth landmarks; 
2) Successful Detection Rate (SDR) is defined as the percentage of landmarks that are accurately detected within a range of 2.0 mm, 2.5 mm, 3.0 mm, and 4 mm from the ground-truth landmarks.

\subsection{Implementation Details}
\label{sec:implementation}
Our CeLDA employs U-Net \cite{ronneberger2015u} as the network backbone $f_{\theta}$.
In Eq.~(\ref{eq:ema}), $\alpha$ = 0.99 and the mini-batch size $|\mathcal{B}|$ = 8.
All images are resized to 512 $\times$ 512 as model input. 
Training images are augmented to introduce random changes in brightness, contrast, and Gaussian noise. 
Our CeLDA is optimized for a total of 150 training epochs, using SGD optimizer with a learning rate of 0.001 which is decreased by a factor of 0.1 per 50 epochs.
$\lambda_1$ and $\lambda_2$ in Eq.~(\ref{eq:totalloss}) are set to 1.0 and 3.0 respectively. We have the mask ratio $R$ = 0.7 for the prototype relation mining.
All methods were implemented in the PyTorch framework and trained on an NVIDIA Tesla A100 GPU with 40GB memory.

\subsection{Comparison with SOTA Approaches}
\input{results}

We compare our CeLDA with the following typical landmark detection models.
Cascade RCNN~\cite{cai2018cascade} detects cephalometric landmarks with a multi-stage object detection architecture to progressively eliminate noisy predictions.  
SCN~\cite{payer2019integrating} regresses landmark heatmaps with a fully convolutional network that considers the spatial configuration of landmarks. 
GU2Net~\cite{zhu2021you} is a universal landmark detection method that solves multiple detection tasks with end-to-end training on mixed datasets.
We also compare with the recent champion method, proposed by Wu et al.~\cite{wu2023revisiting}, in the MICCAI CL-Detection2023 leaderboard. 
It is worth noting that all the above approaches are designed for detecting landmarks from only adult images. 
To achieve comparison fairness, all these competing approaches employ the same image augmentation strategies mentioned in Section~\ref{sec:implementation}.

Table~\ref{tabel:results} provides landmark detection results on CephAdoAdu test set. 
As evident, our CeLDA consistently outperforms other competing approaches on both adult and adolescent cases, only adult cases, and only adolescent cases.
In particular, our CeLDA exhibits more improvements (in both MRE and SDR metrics) for only adolescent cases compared with only adult cases, showing its strength in detecting more challenging adolescent cephalometric landmarks.
Moreover, CeLDA also greatly surpasses other approaches in both adult and adolescent cases, 
verifying its effectiveness in detecting landmarks across age groups.

\subsection{Analytical Ablation Studies}
\input{ablation}

\begin{figure}[t!]
\centering
    \subfigure[]{\includegraphics[width=0.69\textwidth, height=5 cm]{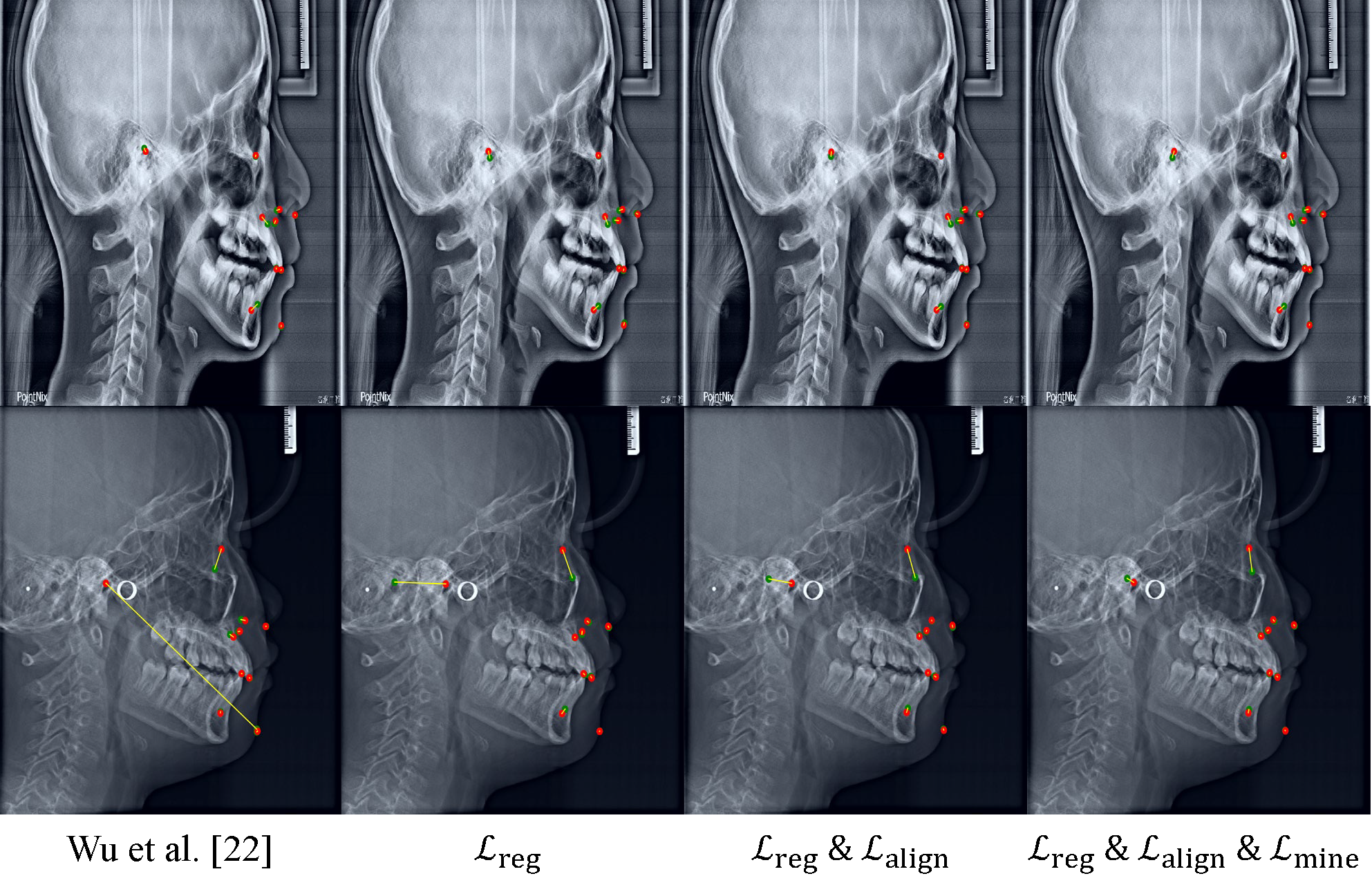}}
    \hfill
    \subfigure[]{\includegraphics[width=0.30\textwidth, height=5 cm]{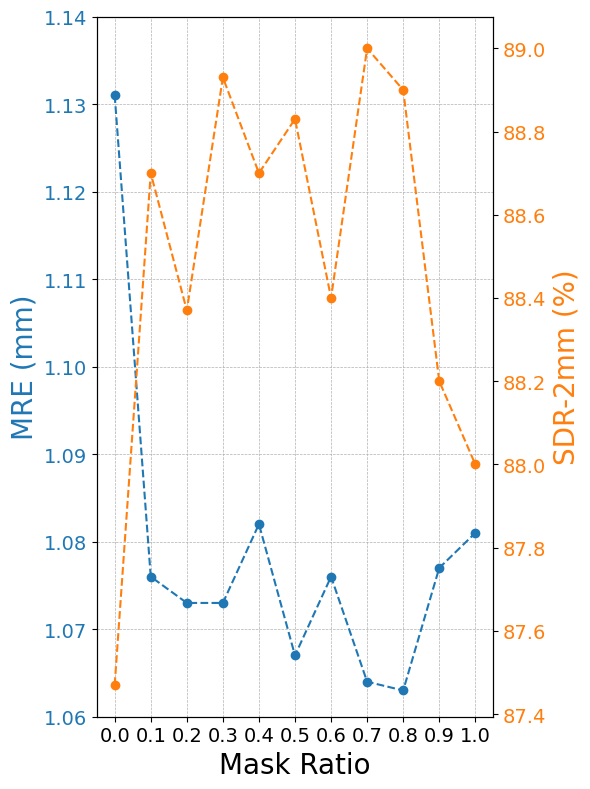}}
    \caption{(a) Visual comparison of different landmark detection methods, where the ground truth and predicted landmarks are indicated in green and red dots respectively, and each pair of them is linked with a yellow line.
    (b) Effect of the mask ratio $R$ for mining landmark prototype relations. }
    \label{fig:ablation}
\end{figure}

We perform ablation experiments to study the effectiveness of our prototype alignment $\mathcal{L}_{\text{align}}$ and prototype relation mining $\mathcal{L}_{\text{mine}}$, with results given in Table~\ref{tabel:ablation}. 
We observe that the baseline (using only $\mathcal{L}_{\text{reg}}$) achieves an MRE of 1.16 mm on adult and adolescent cases, which reduces to 1.13 mm, upon the use of prototype alignment loss. 
Remarkable performance improvements can be observed when further utilizing the prototype relation mining strategy, 
showing the advantage of mining prototype relations to harness the anatomical dependency between landmarks. 
We show typical visual landmark detection results in Fig.~\ref{fig:ablation} (a), 
where we observe progressive improvements with the incorporation of each key component in our method.

In Fig.~\ref{fig:ablation} (b), we explore the sensitivity of our CeLDA to the mask ratio $R$ used for the landmark prototype relation mining. 
As evident, a small mask ratio is inadequate to mine the prototype relations, resulting in sub-optimal results. 
Conversely, a large mask ratio may lead CeLDA to reconstruct wrong landmark prototypes, causing a decline in predictive performance.
According to Fig.~\ref{fig:ablation} (b), we set the mask ratio at 0.7 in all other experiments.

\section{Conclusion}
In this work, we presented the CeLDA method to address cephalometric landmark detection across different age groups with the prototypical network. 
Our CeLDA detects cephalometric landmarks by comparing image features with a set of holistic landmark prototypes, where their anatomical relations are exploited with a masking-based mining strategy. 
Our CeLDA shows great superiority over existing approaches on adolescent and adult cases.
We established and released the first cephalometric benchmark dataset covering a large number of both adult and adolescent cases, with the hope that it will provide a more comprehensive evaluation for the landmark detection community.

%
%
\bibliographystyle{splncs04}
\bibliography{references.bib}

\end{document}


\title{Cephalometric Landmark Detection across Ages with Prototypical Network \\ Supplementary Materials}
%
\maketitle

\begin{figure}
    \centering
    \includegraphics[width=\textwidth]{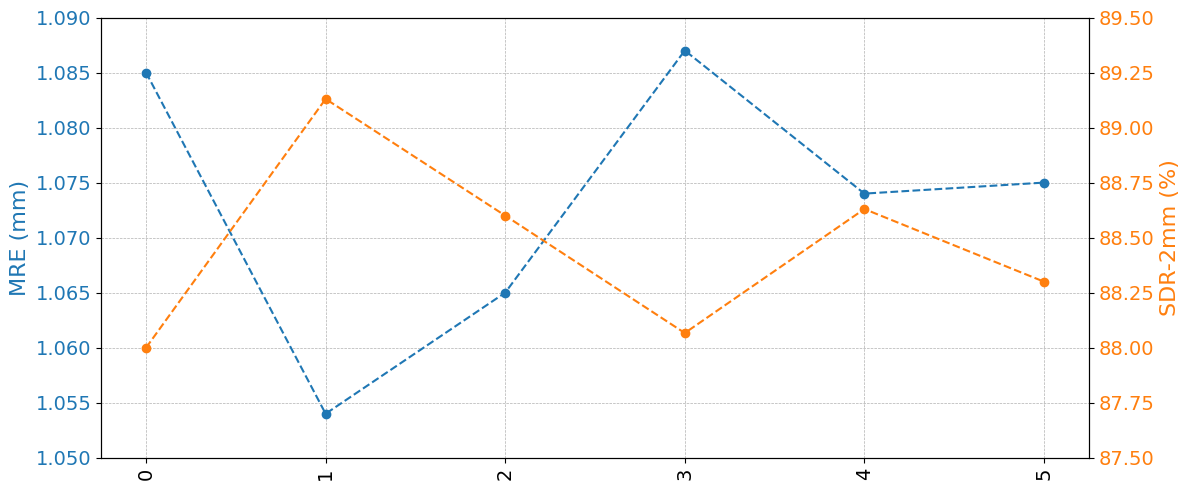}
    \caption{Sensitivity analysis of $\lambda_1$, with $\lambda_2$ fixed as 3.0 }
\end{figure}

\begin{figure}
    \centering
    \includegraphics[width=\textwidth]{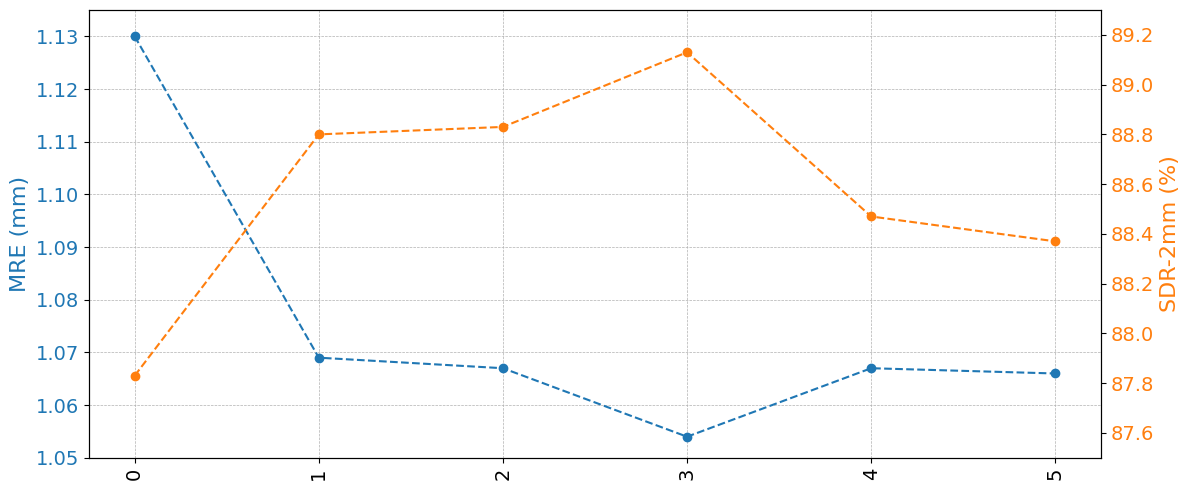}
    \caption{Sensitivity analysis of $\lambda_2$, with $\lambda_1$ fixed as 1.0 }
\end{figure}

%% file: results.tex
\begin{table}[t!]
\caption{Cephalometric landmark detection results with both adult and adolescent cases, only adult cases, and only adolescent cases, respectively.}
\label{tabel:results}
\resizebox{\columnwidth}{!}{%
\begin{tabular}{l|ccccc|ccccc|ccccc}
\hline
\multirow{3}{*}{Methods} &
  \multicolumn{5}{c|}{Adult + Adolescent} &
  \multicolumn{5}{c|}{Adult} &
  \multicolumn{5}{c}{Adolescent} \\ \cline{2-16} 
 &
  \multicolumn{1}{c|}{\multirow{2}{*}{\begin{tabular}[c]{@{}c@{}}MRE $\downarrow$\\ (mm, std.)\end{tabular}}} &
  \multicolumn{4}{c|}{SDR (\%) $\uparrow$} &
  \multicolumn{1}{c|}{\multirow{2}{*}{\begin{tabular}[c]{@{}c@{}}MRE $\downarrow$\\ (mm, std.)\end{tabular}}} &
  \multicolumn{4}{c|}{SDR (\%) $\uparrow$} &
  \multicolumn{1}{c|}{\multirow{2}{*}{\begin{tabular}[c]{@{}c@{}}MRE $\downarrow$\\ (mm, std.)\end{tabular}}} &
  \multicolumn{4}{c}{SDR (\%) $\uparrow$} \\ \cline{3-6} \cline{8-11} \cline{13-16} 
 &
  \multicolumn{1}{c|}{} &
  2mm &
  \multicolumn{1}{c}{2.5mm} &
  \multicolumn{1}{c}{3mm} &
  \multicolumn{1}{c|}{4mm} &
  \multicolumn{1}{c|}{} &
  \multicolumn{1}{c}{2mm} &
  \multicolumn{1}{c}{2.5mm} &
  \multicolumn{1}{c}{3mm} &
  \multicolumn{1}{c|}{4mm} &
  \multicolumn{1}{c|}{} &
  \multicolumn{1}{c}{2mm} &
  \multicolumn{1}{c}{2.5mm} &
  \multicolumn{1}{c}{3mm} &
  \multicolumn{1}{c}{4mm} \\ \hline
Cascade RCNN \cite{cai2018cascade} & 
  \multicolumn{1}{c|}{2.31 (0.94)} & 61.47 & 73.20 & 81.13 & 90.77 &
  \multicolumn{1}{c|}{2.19 (0.97)} & 59.93 & 72.13 & 80.47 & 90.80 &
  \multicolumn{1}{c|}{2.43 (0.94)} & 63.00 & 74.27 & 81.80 & 90.73 \\
SCN \cite{payer2019integrating} &
  \multicolumn{1}{c|}{1.73 (1.06)} & 82.97 & 90.40 & 93.37 & 96.57 &
  \multicolumn{1}{c|}{1.40 (0.48)} & 82.07 & 91.20 & 94.33 & 97.33 &
  \multicolumn{1}{c|}{2.05 (1.70)} & 83.87 & 89.60 & 92.40 & 95.80 \\
GU2Net \cite{zhu2021you} &
  \multicolumn{1}{c|}{1.69 (0.91)} & 80.33 & 88.13 & 91.47 & 95.57 &
  \multicolumn{1}{c|}{1.46 (0.50)} & 80.27 & 88.80 & 92.07 & 96.33 &
  \multicolumn{1}{c|}{1.93 (1.35)} & 80.40 & 87.47 & 90.87 & 94.80 \\
Wu et al. \cite{wu2023revisiting} &
  \multicolumn{1}{c|}{1.34 (1.24)} & 87.17 & 91.93 & 95.57 & 97.10 &
  \multicolumn{1}{c|}{1.13 (0.66)} & 86.60 & 92.13 & 95.00 & 97.80 &
  \multicolumn{1}{c|}{1.55 (1.87)} & 87.73 & 91.73 & 94.13 & 96.40 \\
CeLDA &    
 \multicolumn{1}{c|}{\textbf{1.05 (0.33)}} & \textbf{89.13} & \textbf{93.60} & \textbf{96.17} & \textbf{98.67} &
\multicolumn{1}{c|}{\textbf{1.10 (0.37)}} & \textbf{88.33} & \textbf{92.93} & \textbf{96.20} & \textbf{98.80} &
\multicolumn{1}{c|}{\textbf{1.00 (0.34)}} & \textbf{89.93} & \textbf{94.27} & \textbf{96.13} & \textbf{98.53} \\ \hline

\end{tabular}%
}
\end{table}

%% file: ablation.tex
\begin{table}[t]
\caption{Ablation analysis for our proposed CeLDA method. }
\label{tabel:ablation}
\resizebox{\columnwidth}{!}{%
\begin{tabular}{ccc|ccccc|ccccc|ccccc}
\hline
\multirow{3}{*}{$\mathcal{L}_{\text{reg}}$} &
  \multirow{3}{*}{$\mathcal{L}_{\text{align}}$} &
  \multirow{3}{*}{$\mathcal{L}_{\text{mine}}$} &
  \multicolumn{5}{c|}{Adult + Adolescent} &
  \multicolumn{5}{c|}{Adult} &
  \multicolumn{5}{c}{Adolescent} \\ \cline{4-18} 
 &
   &
   &
  \multicolumn{1}{c|}{\multirow{2}{*}{\begin{tabular}[c]{@{}c@{}}MRE $\downarrow$\\ (mm, std.)\end{tabular}}} &
  \multicolumn{4}{c|}{SDR (\%) $\uparrow$} &
  \multicolumn{1}{c|}{\multirow{2}{*}{\begin{tabular}[c]{@{}c@{}}MRE $\downarrow$\\ (mm, std.)\end{tabular}}} &
  \multicolumn{4}{c|}{SDR (\%) $\uparrow$} &
  \multicolumn{1}{c|}{\multirow{2}{*}{\begin{tabular}[c]{@{}c@{}}MRE $\downarrow$\\ (mm, std.)\end{tabular}}} &
  \multicolumn{4}{c}{SDR (\%) $\uparrow$} \\ \cline{5-8} \cline{10-13} \cline{15-18} 
 &
   &
   &
  \multicolumn{1}{c|}{} &
  2mm &
  2.5mm &
  3mm &
  4mm &
  \multicolumn{1}{c|}{} &
  2mm &
  2.5mm &
  3mm &
  4mm &
  \multicolumn{1}{c|}{} &
  2mm &
  2.5mm &
  3mm &
  4mm \\ \hline
$\checkmark$  & 
&
   &
  \multicolumn{1}{c|}{1.16 (0.38)} & 86.77 & 92.00 & 95.30 & 98.10 &
\multicolumn{1}{c|}{1.16 (0.36)} & 85.93 & 91.73 & 95.60 & 98.53 &
\multicolumn{1}{c|}{1.17 (0.42)} & 87.60 & 92.27 & 95.00 & 97.67 \\
$\checkmark$  &
  $\checkmark$  &
   &
  \multicolumn{1}{c|}{1.13 (0.34)} & 87.83 & 92.97 & 95.70 & 98.30 &
\multicolumn{1}{c|}{1.14 (0.36)} & 86.20 & 92.47 & 96.07 & 98.87 &
\multicolumn{1}{c|}{1.12 (0.44)} & 89.47 & 93.47 & 95.33 & 97.73 \\
$\checkmark$ &
  $\checkmark$ &
  $\checkmark$ &
  \multicolumn{1}{c|}{{1.05 (0.33)}} & {89.13} & {93.60} & {96.17} & {98.67} &
\multicolumn{1}{c|}{{1.10 (0.37)}} & {88.33} & {92.93} & {96.20} & {98.80} &
\multicolumn{1}{c|}{{1.00 (0.34)}} & {89.93} & {94.27} & {96.13} & {98.53} \\
 \hline
\end{tabular}%
}
\end{table}